\RequirePackage{fix-cm}

\documentclass[smallextended]{svjour3}       % onecolumn (second format)

\smartqed  % flush right qed marks, e.g. at end of proof

\usepackage{graphicx}
\usepackage{amsmath}
\usepackage{stfloats}
\usepackage{subfig}
\usepackage{float}

\begin{document}

\title{Patient-Specific 3D Volumetric Reconstruction of Bioresorbable Stents}

\subtitle{A Method to Generate 3D Geometries for Computational Analysis of Coronaries Treated with Bioresorbable Stents}

\author{Boyi~Yang \and Marina~Piccinelli \and Gaetano~Esposito \and Tianli~Han \and Yasir~Bouchi \and
	Bill~Gogas \and Don~Giddens \and Habib~Samady \and Alessandro~Veneziani}

\institute{	B. Yang  \at
	Department of Biomedical Informatics, Harvard Medical School, Harvard University \\    
	10 Shattuck St, Boston, MA 02115 \\
	\and
	B. Yang \and G. Esposito \and  T. Han \and A. Veneziani \at
	Department of Mathematics and Computer Science, Emory University \\    
	\and
	M. Piccinelli \at
	Department of Radiology, Emory University Hospital
	\and 
	B. Gogas \and D. Giddens \and H. Samady \at
	Division of Cardiology, Emory University Hospital \\
}
\date{}
%\date{Received on December 16th, 2016 / Revised on September 26th, 2018}

\maketitle

\begin{abstract}
As experts continue to debate the optimal surgery practice for coronary disease – percutaneous coronary intervention (PCI) or coronary aortic bypass graft (CABG) – computational tools may provide a quantitative assessment of each option. Computational fluid dynamics (CFD) has been used to assess the interplay between hemodynamics and stent struts; it is of particular interest in Bioresorbable Vascular Stents (BVS), since their thicker struts may result in impacted flow patterns and possible pathological consequences. Many proofs of concept are presented in the literature; however, a practical method for extracting patient-specific stented coronary artery geometries from images over a large number of patients remains an open problem. 

This work provides a possible pipeline for the reconstruction of the BVS. Using Optical Coherence Tomographies (OCT) and Invasive Coronary Angiographies (ICA), we can reconstruct the 3D geometry of deployed BVS in vivo. We illustrate the stent reconstruction process: (i) automatic strut detection, (ii) identification of stent components, (iii) 3D registration of stent curvature, and (iv) final stent volume reconstruction. The methodology is designed for use on clinical OCT images, as opposed to approaches that relied on a small number of virtually deployed stents.

The proposed reconstruction process is validated with a virtual phantom stent, providing quantitative assessment of the methodology, and with selected clinical cases, confirming feasibility. Using multimodality image analysis, we obtain reliable reconstructions within a reasonable timeframe. This work is the first step toward a fully automated reconstruction and simulation procedure aiming at an extensive quantitative analysis of the impact of BVS struts on hemodynamics via CFD in clinical trials, going beyond the proof-of-concept stage.
	
	
\end{abstract}

\section{Introduction}
\label{intro}
Percutaneous coronary intervention (PCI),  one of the most commonly performed procedure for the treatment of coronary artery disease (CAD), re-opens occluded vessel by a coronary stent at the location of the atherosclerotic plaque. Different types of stents have been developed to improve clinical outcomes. The Bioresorbable Vascular Stent (BVS) is a new generation of stent introduced to the clinical practice \cite{ormiston2008bioabsorbable}. The BVS systems work similarly to traditional metallic stents but are composed of a material that can be absorbed in about three years, leaving no permanent scaffold in the treated artery. To stand the stress during and after deployment, the new BVS demands thicker struts than metallic devices. For instance, the strut thickness of the Abbott Xience Metallic stent is $91\ \mu m$, while the thickness of Abbott Absorb BVS is $150\ \mu m$ (Abbott Vascular, Santa Clara, CA). The larger strut thickness of BVS may trigger a nontrivial interplay between the blood dynamics and the vessel wall.  A direct relationship between the blood flow characteristic and atherosclerosis mechanisms is widely accepted: areas of disturbed and oscillating flow are prone to plaque progression (see, e.g., \cite{cunningham2005role,chatzizisis2007role,malek1999hemodynamic}). Pre-clinical and clinical evidence on the usage of stents in general and BVS in particular \cite{koskinas2012role,carlier2003blommerde,ladisa2005alterations,chen2009effects,zhang2013bioresorbable} suggest that local flow patterns are likely to also be responsible for complications such as restenosis, neointimal hyperplasia, and stent thrombosis. The stent introduces acute changes in the vessel anatomy. The struts protrusion against the vessel wall leads to focal geometric irregularities, which create the disturbances in the laminar flow patterns and in the vessel wall biology indicated as specific atherogenic stimuli \cite{mejia2009evaluation,wentzel2001relationship,sanmartin2006influence,gijsen2003usefulness}. 

Computational modeling is the most appropriate tool to predict the potential impact of anomalous flow patterns on the outcome of the therapy. In particular, Computational Fluid Dynamics (CFD) for the study of cardiovascular diseases has been developed, validated, and applied to a variety of patient-specific anatomies and vascular districts \cite{formaggia2010cardiovascular,steinman2002image,taylor2010image}. The CFD approach requires an accurate geometrical reconstruction of the 3D vessel lumen with the presence of the stent struts evident in OCT. This work accomplishes such geometrical reconstruction, thus enabling in-depth CFD analysis at the strut level of fully patient-specific models.

Optical Coherence Tomography (OCT), with its high resolution, allows reliable detection of the struts and has been used in clinical trials that investigate PCI outcomes to assess the rate of BVS absorption and to inspect the response of the vessel wall to the stent \cite{gonzalo2009optical}. Because of the translucent polymer of BVS, OCT is also particularly suitable for BVS imaging \cite{sheehy2012vivo,wang2014automatic}. 

The state of the art of computational modeling of stented arteries is summarized in \cite{morlacchi2013modeling}. The challenges of OCT and multimodal imaging are identified. Early computational studies with stented arteries were based on IVUS-reconstructed geometries with no stent appearance \cite{sanmartin2006influence} or on idealized stent models \cite{jimenez2009hemodynamically}. More recently, manual segmentation of OCT images enabled steady simulations of flow \cite{papafaklis2013vivo} even though the manual procedure - of which reliability is hard to assess - seems to be quite operator-dependent and not prone to the automation required by a large clinical trial. An improvement of this work within the same guidelines is presented in \cite{bourantas2014effect}. Another approach to reconstruct stented vessel is to combine the patient-specific lumen geometry from angiography with a virtually deployed stent. Virtual stent deployment is achieved using a series of Boolean operations \cite{ellwein2011optical}, or by mechanical simulation of balloon expansion \cite{chiastra2015computational}. The latter work presents results on images obtained by OCT and CT on two clinical cases, with an estimated maximal error of 20.4 \% on the area of the reconstructed slice.

A direct segmentation approach for extracting the segmented artery was attempted in \cite{gogas2013biomechanical,yang2015novel,gogas2016novel} by our team, and in \cite{bourantas2014fusion}. Recent advancements on OCT-derived patient-specific metallic stent reconstruction is presented in \cite{o2016constraining}. In this case, the OCT images are processed and the reconstruction is guided by an educated combination of {\it a priori} information on the stent design. Their approach is presented on one non-human (porcine) case.

In this paper, we present a methodology to reconstruct the patient-specific 3D BVS stent geometry from a combination of 2D OCT and ICA images on real patients. The 3D stent geometry is the basis of the reconstruction of the stented lumen for future work on CFD analysis. The entire stent reconstruction method was applied to (i) a synthetic computerized phantom of an idealized stent under different deformations and (ii) a group of patients enrolled in the ABSORB Clinical Study at the Emory Cardiovascular Imaging \& Biomechanics Core Laboratory. The former validates the methodology and identifies the primary source of errors; the latter demonstrates that the method works on real patients and provides an effective tool to quantitatively investigate BVS in clinical scenarios. Our ultimate goal is to deliver a computational framework to be used in clinical practice. Automation (or semi-automation), robustness and reliability are the requirements needed by the procedure. To the best of our knowledge, this is the first methodological paper presented on a multi-patient ($>10$) study. 

\section{Materials}
The stent reconstruction method was developed using clinical OCT images. Then we validated the method over a 3D virtual stent geometry.

\subsection {Clinical OCT Images}
A database of clinical OCT acquisitions performed after the deployment of Abbott bioresorbable stents with corresponding baseline bi-plane coronary angiographies is available at the Emory Cardiovascular Imaging \& Biomechanics Core Laboratory within the imaging sub-study of the ABSORB III Clinical Trial \cite{gogas2013biomechanical}. We have successfully applied the proposed methodology and reconstructed 3D stents for sixteen cases ($N =16$) each of which has both the OCT images for stent detection and the angiography for curvature extraction. We selected four cases (Case 2, 3, 6, and 12) for demonstration and validation as these cases are representative of the general characteristics of the results. The treated artery was the right coronary artery (RCA) for Case 2, 3 and 12 and the left circumflex artery (LCX) for Case 6.  The OCT scan was performed according to the standard clinical protocols. Each OCT frame is sized $1024\times1024$ pixels with an in-plane spatial resolution that was 0.009 mm for Case 2 and 0.007 mm for Case 3, 6 and 12. The inter-frame thickness was imported from image file headers and ranged from 0.1 to 0.2 mm. The length of the OCT acquisition varied among cases. In addition, the stent length also varies from patient to patient depending on the size of the plaque. Hence the number of OCT frames that show stent struts ranged from 63 frames to 149 in the four selected cases. The baseline angiographies were acquired at different instants throughout the procedure according to standard clinical protocols. The angiographic series depicting the catheter wire positioned inside the vessel being treated were selected to retrieve the 3D anatomy of the artery.

\subsection {A Virtual Stent Geometry for Validation}
\begin{figure*}[htbp]
	\centering
	\includegraphics[width=4.7in]{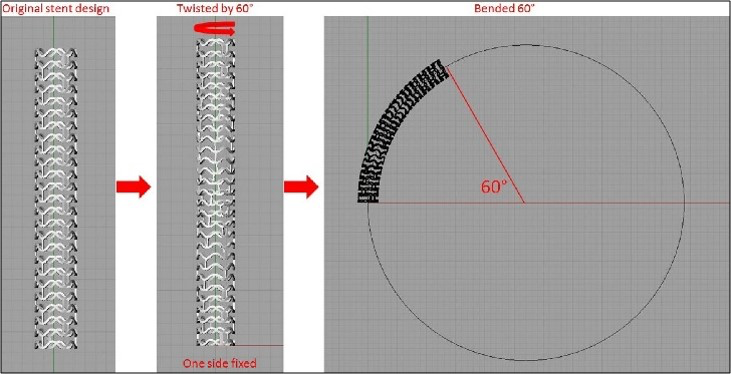}
	\caption{A virtual phantom stent geometry constructed from the stent design.  
		(1) Stent twisted by 60$^\circ{}$. (2) Stent bended along a 60$^\circ{}$ circular sector. }
	\label{fig_1}
\end{figure*}

A virtual phantom of a deployed coronary stent was created for validation. The original stent design STL geometry is obtained from Abbott Laboratory. This geometry is originally in straight and un-deployed configuration. To mimic its realistic deployed state, we first twisted it along its axis by 60$^\circ$ and then bent it to match the 60$^\circ$ circular section depicted in Fig. \ref{fig_1}. This operation is intended to reproduce the distortions of the structure during the deployment. Both operations on the BVS design were performed by Rhinoceros 5.0 (McNeel North America, Seattle, WA 98103). To mimic the OCT acquisition of the virtual phantom, we virtually sliced the stent perpendicularly to its centerline at the intervals of 0.1 mm and 0.2 mm respectively, for two benchmarks, to quantify the impact of the slice resolution. The profiles resulting from this slicing were used to create binary images. The presence of the shadow produced in the real images by the wire tip was added by masking a sector of the binary images that smoothly change location throughout the image sequence. The purely binary nature of the virtual OCT frames facilitated the detection of the struts as opposed to the real colored OCT images. Nonetheless, the ultimate purpose is to test the ability of the proposed reconstruction methodology on a reliable 3D structure after the comparison with the phantom stent that represents the exact solution. In result, we give a quantitative assessment of the procedure as a function of the axial resolution.

\section{Method}
Our patient-specific stent reconstruction process can be divided into four steps: (i) Strut Detection, (ii) Strut Point Connectivity Recovery, (iii) Angiography Registration, and (iv) Stent Volumetric Reconstruction.

\subsection{Strut Detection}
\label{Strut Detection}
The strut detection procedure consists of four steps: (i) Region of Interest (ROI) Identification, (ii) Iterative Strut Detection, (iii) False Positive Filtering, and (iv) Correction and Patching. All the image processing tasks were performed with Matlab R2016a (MathWorks Inc., Natick, MA, 01760). The details of the Matlab functions used for detection can be found in \cite{graythresh}.

\subsubsection{Finding the Region of Interest (ROI)}
\begin{figure*}[htbp]
	\centering
	\includegraphics[width=4.7in]{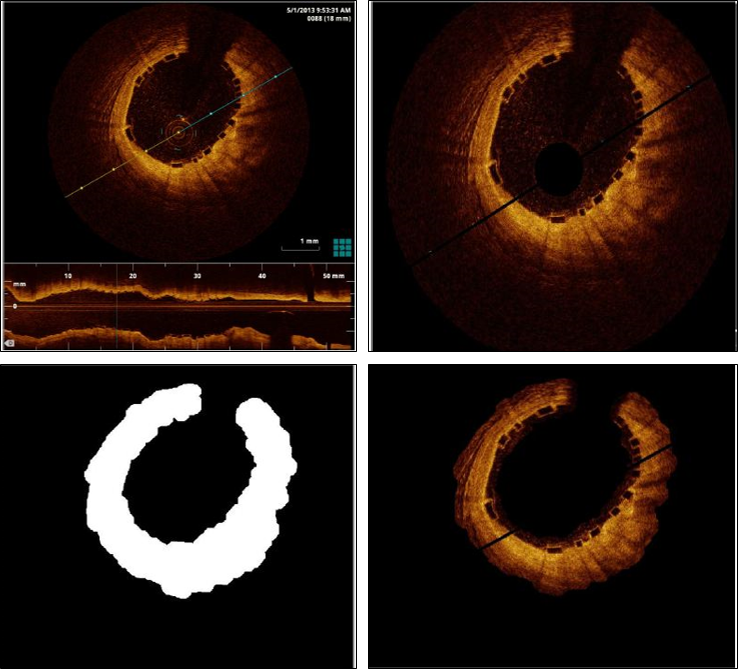}
	\caption{Top Left: A baseline OCT image. The lower part of the image is the longitudinal review of the stented vessel. 
		Top Right: The OCT image after the unnecessary features are removed automatically. Bottom Left: The binary mask for the extraction of the ROI. 
		Bottom Right: The ROI for strut detection. }
	\label{fig_1r}
\end{figure*}

Our region of interest (ROI) is the neighborhood of the vessel wall that appears in high-intensity values. After removing the unnecessary features of the OCT images (as shown in Fig. \ref{fig_1r}, a typical OCT image contains features that are generally disturbing to the strut detection) to reduce the image to the ROI, we first convert the original RBG OCT image to grayscale image. We find the threshold of the grayscale image by Otsu's method \cite{otsu1975threshold}, then convert the OCT to a binary image. Otsu method finds the optimal thresholding by maximizing the inter-class variance between the two classes of pixels into which the image is split. The vessel wall area has higher intensities values, so after thresholding this area becomes white, and we name it $A_V$. The rest of the OCT is turned to black. This step is performed by the Matlab function {\tt graythresh}. To avoid loss of important details, such as the malapposed struts, $A_V$ is then further expanded to $A_{ROI}$ by an appropriate number of pixels: $A_{ROI}$ includes $A_V$ and an additional distance to safely include all the struts. This expansion is performed by the Matlab erosion function {\tt imerode} which enlarges the white area (Fig. \ref{fig_1r}). This final white area is used to mask the original image in order to obtain the ROI.

\subsubsection{Iterative Primary Strut Detection}
\begin{figure*}[htbp]
	\centering
	\subfloat[]{\includegraphics[width=3.5in,height=1.7in]{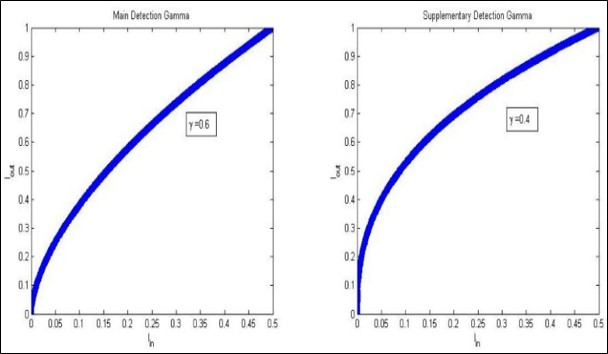}%
		\label{fig_2l}}
	\hfil
	\subfloat[]{\includegraphics[width=3.5in,height=1.7in]{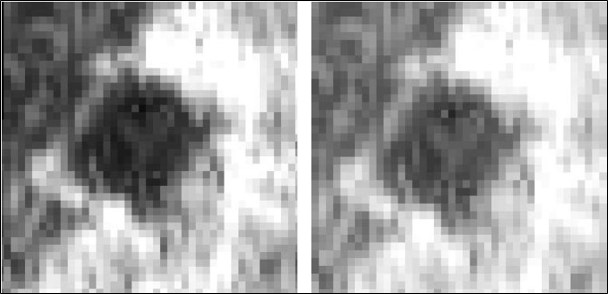}%
		\label{fig_2c}}
	\hfil
	\subfloat[]{\includegraphics[width=3.5in,height=1.7in]{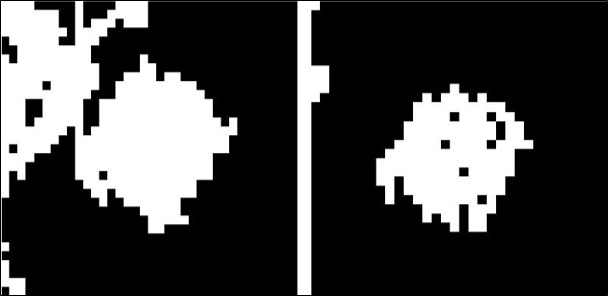}%
		\label{fig_2r}}
	\caption{(a) Gamma correction function curve with $\gamma = 0.6$ (left) and $\gamma= 0.4$ (right). 
		(b) The grayscale image of a strut after gamma correction with $\gamma = 0.6$ (left) and $\gamma= 0.4$ (right). 
		(c) Binary image of the strut after using $\gamma = 0.6$ (left) $\gamma= 0.4$ (right). Shown on the left, the strut is missed by the primary detection since the white area is not fully closed. Shown on the right, the same strut is captured by the second iteration.}
	\label{fig_2}
\end{figure*}

To emphasize the details relevant to the strut detection, we applied a Gamma Function correction to the intensity values of the grayscale ROI \cite{poynton2012digital}. The grayscale is mapped from a low intensity interval $[a,b]$ to a high-intensity interval $[c,d]$ using the Gamma correction function
$$
I_{out}  = (d-c) \left(\dfrac{I_{in}-a}{b-a}\right)^\gamma+c, 
$$
where $I_{in} \in [a,b]$ and 
$I_{out}\in [c,d]$.  
We choose $[c, d] = [0, 1]$ and $a=0$. The value $b$ is selected using the intensity histogram of the grayscale ROI. We use Matlab function {\tt imhist} to divide the intensity value into 256 levels and count the number of pixels in each level. We set the threshold of pixel count to be 20 and search for the first intensity level $i$ with less than 20 pixels, then $b=i/256$. Since $i$ is usually around 155,  $b$ is usually around 0.6. The values of $\gamma$ drive the warping of the gray level (Fig. \ref{fig_2l}). When $\gamma=1$, the mapping function sends one interval linearly onto another. For $\gamma<1$, the mapping function is weighted toward high-intensity values, so that the function brings more dark pixels from low-intensity to high-intensity values. The close-up gamma-corrected images of the same strut with $\gamma=0.6$ and $\gamma=0.4$ respectively are shown in Fig. \ref{fig_2c}, where we can see that lower gamma leads to an overall brighter grayscale image. The Gamma value is lowered along the iterations of the detection. 

After the correction, we apply Otsu's Method to define an optimal threshold to transform the corrected ROI into a binary image \cite{otsu1975threshold}. Then we apply a binary inversion so that the stent struts appear as a set of white boxes surrounded by black pixels (Fig. \ref{fig_2r}). The actual detection of the struts is then performed by first identifying all the possible candidates. A recursive flood-fill algorithm is applied to detect all regions formed by the connected white pixels. The operation is performed by Matlab function {\tt imfill}. Any enclosed white region is recognized as a stent strut. However, small white regions with less than 50 pixels are immediately discarded because struts should have approximately 300 pixels in the baseline OCT images. The remaining white regions are considered candidate struts. The borders are recorded (the green contour in Fig. \ref{fig_3l}), and the centroids of the candidate regions (the white dots in Fig. \ref{fig_3r}) are stored.

\subsubsection{False Positives Filtering}
To eliminate the candidate struts obtained during the primary detection that are artifacts (false positives), we apply a sequence of filtering actions.

\begin{figure*}[htbp]
	\centering
	\subfloat[]{\includegraphics[width=3.5in,height=1.2in]{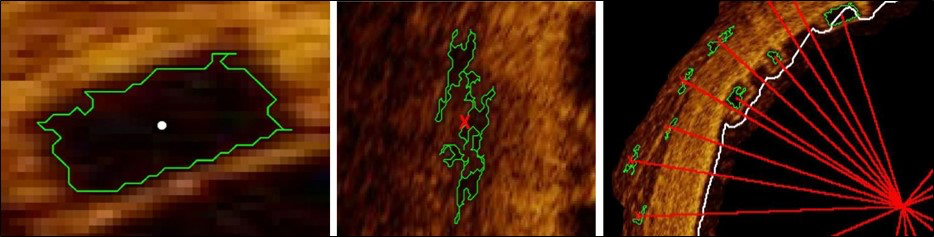}%
		\label{fig_3l}}
	\hfil
	\subfloat[]{\includegraphics[width=3.5in,height=1.7in]{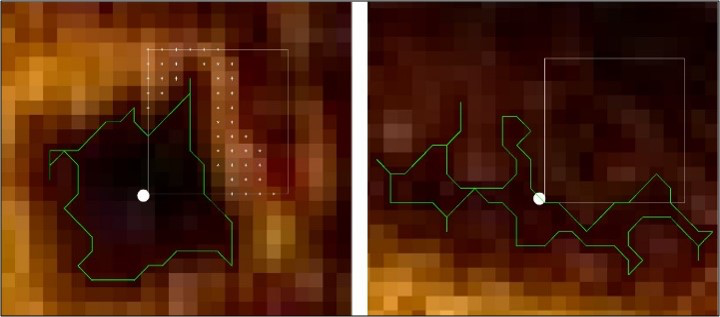}%
		\label{fig_3r}}
	\caption{(a) Left: A detected strut. The green line is the strut contour; the white dot is the strut centroid. 
		Center: A candidate strut eliminated by the Contour length filter. Right: Several candidate struts eliminated by the Wall distance filter. (b) Left: A strut that passed the Pixel counter filter with 44 high intensity pixels in the probing square. Right: A strut eliminated by the Pixel count filter. The scan square contains no pixel with high-intensity value.}
	\label{fig_3}
\end{figure*}

\paragraph{Contour Length Filter}
The strut is 150 $\mu m$ in length along the centerline and 200 $\mu m$ transversally at the baseline, corresponding to approximately 20 pixels. Therefore, the perimeter of a strut box should be at least 80 pixels long if the strut is in a rectangular shape. A candidate strut contour longer than 250 pixels or shorter than 20 is considered a false positive and it is eliminated from the candidates (see Fig. \ref{fig_3l}).

\paragraph {Wall Distance Filter} After deployment, the struts are usually apposed to the lumen boundary. Therefore, most struts should be near the inner vessel wall in the OCT image (the white line in Fig. \ref{fig_3l}). Based on the relative distance to the lumen contour, additional false positive are removed. The candidate strut centroid and the wire tip determine a straight line (red lines), which intersects the vessel wall. The distance between this intersection and the candidate strut centroid is measured. If this distance is greater than 40 pixels, the candidate strut is eliminated as a false positive. This filter may be adjusted in case of evident stent malapposition. To avoid redundancy, the struts detected by the primary detection are removed from the grayscale ROI before the supplementary detection process begins; in practice, previously detected struts are simply converted to black so that they will not be re-detected in subsequent detection iterations. We iterate the detection-filtration-removal cycle introduced above and lower the gamma value after each iteration. This procedure is largely based on an empirical tuning obtained after a trial and error training.  The brightness of strut edges varies, and we can catch the maximum amount of struts by applying a series of different $\gamma$ values. We found effective to perform four iterations with the sequence of $\gamma$ values $\gamma_1=0.6, \gamma_2=0.5 , \gamma_3=0.4$, and $\gamma_4=0.35$.

\paragraph {Pixel Count Filter} Another way of detecting false positives is to check the shape of the candidate strut regions. In fact, we filter the candidates featuring irregular regions (see Fig. \ref{fig_3r}). For every strut contour candidate, we define a probing square region of $10\times 10$ pixels with the candidate centroid located on one of its corners. This square explores the neighborhood pixels towards the lumen center.  We mark the candidate as a false positive if the number of bright pixels in this region is below a cut-off percentage. The rationale is that a true strut should have a significant portion of the boundary (bright pixels) in the probing region. We use Otsu's threshold obtained during the first detection with $\gamma=0.6$ as the intensity value threshold and a pixel count of 10\% of the pixels in the probing square. Fig. \ref{fig_3r} shows a strut that passed the Pixel counter filter with 44 high-intensity pixels in the probing square (left panel)  and a strut eliminated by the Pixel count filter with zero high-intensity pixels in the probing square (right panel).

\paragraph{Manual Correction and Patching} After the above detecting and filtering procedure we can automatically detect 80\% to 85\% of the struts, depending on the quality of the images. Such accuracy is sufficient for a qualitative assessment of stent deployment. However, for our ultimate goal of 3D stent reconstruction, we aim at recording 100\% of struts centroids. Therefore, we incorporate a manual correction and patching step to complete the detection task. An efficient graphical user interface was created to allow the user to control each slice - the user can include missed struts and remove persistent false positive by merely clicking on the image. More advanced noise filtering using machine learning techniques is currently under investigation to reduce the need for manual intervention. 

Another filter we could apply is the {\it eccentricity control filter}. Let {\it eccentricity} be the ratio between the maximal and the minimal distance of the border to the centroid. Since the struts are rectangular, regions featuring an eccentricity greater than a given value can be marked as false positives. However, this filter was not applied to cases we present, as we found most of the eccentricity was already detected by the {\it Contour Length Filter}. 

\subsection{Strut Point Connectivity Recovery}
\begin{figure*}[htbp]
	\centering
	\includegraphics[width=4.7in]{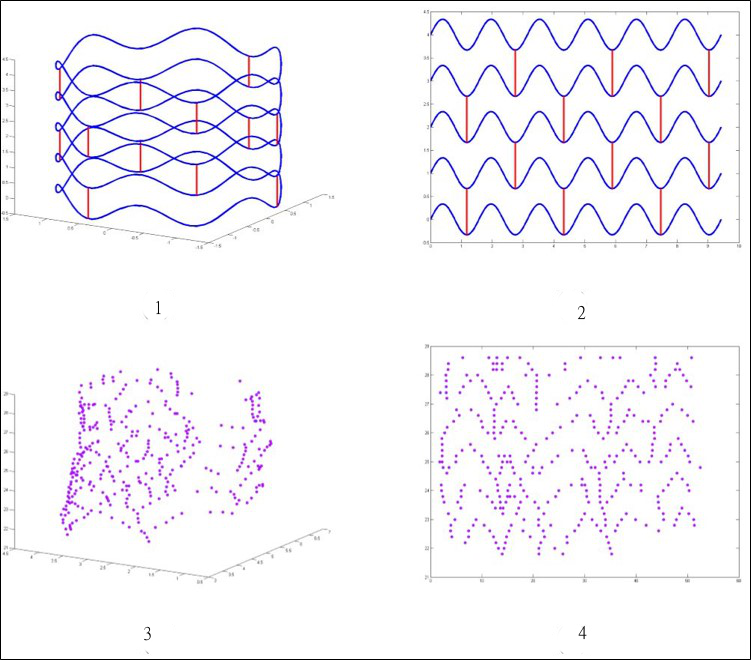}
	\caption{Patterns: (1) 3D stent pattern. (2) 2D stent pattern after unrolling. (3) 3D strut point cloud. (4) 2D strut point cloud for wireframe reconstruction.}
	\label{fig_5l}
\end{figure*}

An undeployed BVS in our dataset features two basic topological components: {\it ring}, i.e. the closed ring-like structure with a sinusoidal shape shown as the blue line in Fig. \ref{fig_5l}; {\it beam}, i.e. the axial structure that connects two adjacent rings. The pattern is evident when it is unrolled from the cylindrical structure into a 2D space (Fig. \ref{fig_5l}). To recover the connectivity of the deployed stent, we need to classify each point as part of one of the two components. We perform this task with an interactive approach. Since it is easier for a human operator to identify the structures on a 2D surface, we first map the 3D points onto a plane.

\subsubsection{Flattening to a 2D Region}
Given the spatial image resolution and the inter-slice distance, the strut coordinates can be stored in physical 3D dimensions. Then, we compute the center of mass O of the 3D cloud of points, and for each strut point we calculate the corresponding polar coordinates in a frame of reference centered in O. All points are then sorted according to the angle $\vartheta$ and displayed over the plane $(r\vartheta, z)$. In Fig. \ref{fig_5r}, we illustrate the original 3D point cloud from the OCT of one of the analyzed cases and the flattened results in 2D. It is now evident how the different points belong to each structure and they can be readily classified by the user.

\subsubsection{Interactive Pattern Interpretation}
\begin{figure*}[htbp]
	\centering
	\includegraphics[width=4.7in]{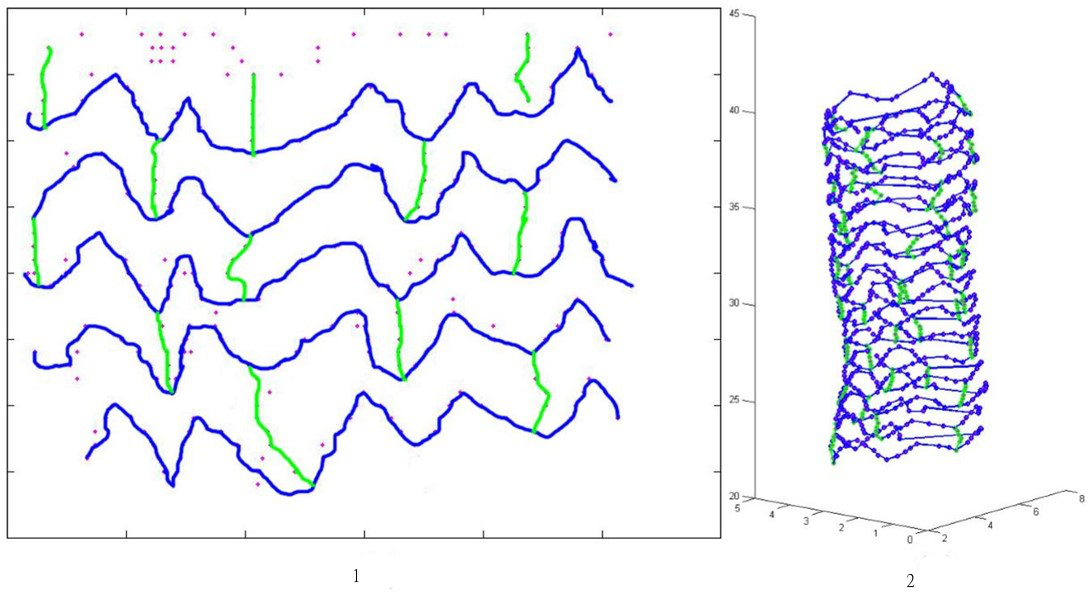}
	\caption{Parsing the cloud: (1) Interpretive drawings by operator. (2) Connected 3D point cloud.}
	\label{fig_5r}
\end{figure*}

An algorithm is developed that allows the operator to interpret the pattern of the 2D point cloud by drawing lines on the plot.  The operator sketches a line that passes the centroids belong to the same ring or beam. The ring lines and beam lines are drawn on the 2D plot by using any image editor with layering feature (e.g. GIMP, Adobe Photoshop). Lines approximating rings are blue while lines approximating beams are green as shown in Fig. \ref{fig_5r}. For each strut point a circular search region is defined to detect the closest ring or beam line, and then the point inherits the line classification. Points that belong to both a ring and beam line are marked as junctions between the two structures (Fig. \ref{fig_5r}). An in-house Matlab application for pattern interpretation was developed to facilitate the operator identification. The operator can categorize a ring or a beam by clicking a series of points directly on the 2D Matlab plot shown in Fig. \ref{fig_5l} Bottom Right.  The closest strut point to each clicked point will be grouped into one ring or beam in a similar fashion as for the sketching method. The point classification then is passed to the 3D point cloud. The categorized 3D point cloud together with the wire path retrieval described in the following section provides the basis for the construction of the stent skeleton. The identification step can be alternatively developed based on procedures that take advantage of the knowledge from the undeployed stent in the same fashion as in \cite{ellwein2011optical,o2016constraining,ladisa2005alterations}. This option is currently under development in the present pipeline, to accelerate the reconstruction procedure and is one of the follow-ups of the present work \cite{LRYV}.

\subsection{Angiography Registration}
\begin{figure*}[htbp]
	\centering
	\includegraphics[width=4.7in]{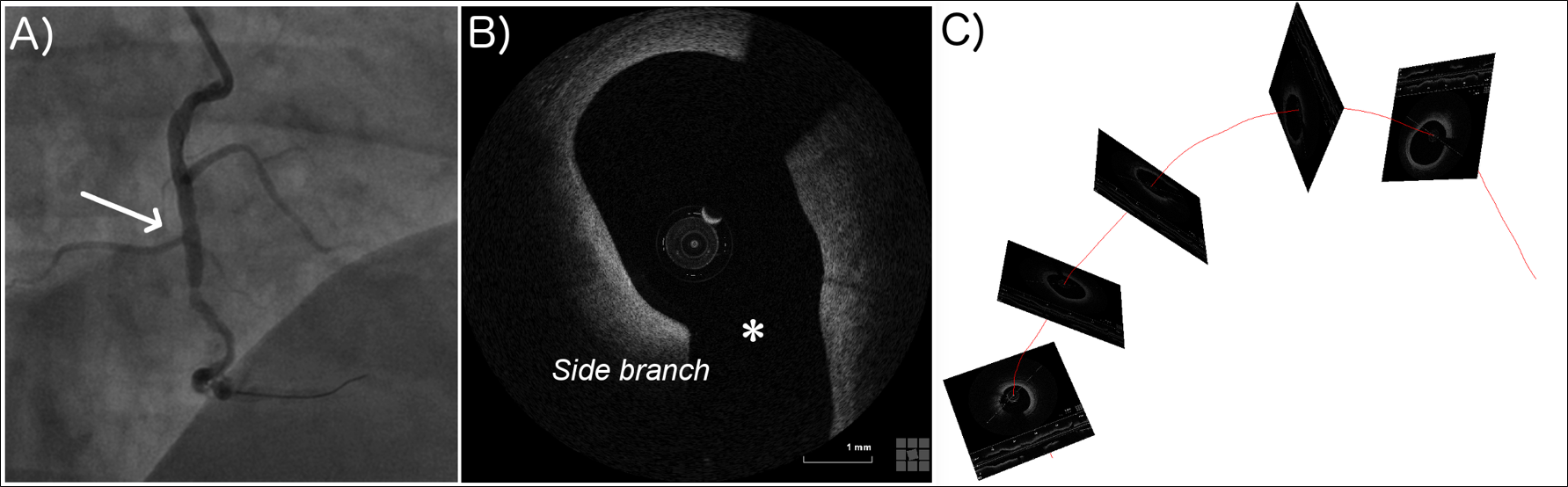}
	\caption{(a) Angiographic view depicting the treated artery and its side branches for one of the analyzed cases; the arrow indicates the branch used to register the angiography to the OCT pullback. 
		(b) OCT frame corresponding to side branch in (a); (c) registration of the OCT frames on top of the OCT wire 3D trajectory. }
	\label{fig_6}
\end{figure*}

Being acquired from inside the vessel, the OCT alone does not allow the retrieval of the true curvature geometry of the vessel. On the other hand, using bi-plane coronary angiography or multiple angiographic views of the same vessel, a 3D reconstruction of the artery being investigated is possible. A well-established and widely used methodology to perform 3D registration of invasive vascular images with ICA was here employed \cite{slager2000true,bourantas2005method}. The 3D OCT catheter pathway within the vessel is initially reconstructed with QAngio XA (Medis Medical Imaging Systems, Leiden, The Netherlands) from different ICA projections. OCT frames are successively positioned along the reconstructed wire by matching each point to the visible wire tip on the corresponding OCT images and by positioning each frame perpendicularly to the wireline. A common point of reference is needed as a starting point to register all the images at the correct location along the reconstructed wire trajectory. This landmark is usually a bifurcation that can be identified on both angiographies and OCT stack. In Fig \ref{fig_6}a one angiographic view of the wire for one of the analyzed case is displayed: it clearly reveals the location of a side branch and allows to measure the distance in mm along the wire. The same side branch can be identified on a specific OCT frame (Fig. \ref{fig_6}b). Once this correspondence has been established all frames can be positioned along the wire - before and/or after the identified landmark with the inter-slice distance obtained from the pullback information (Fig. \ref{fig_6}c). The registration of each image is performed by computing the so-called pointwise Frenet Frame of the line representing the 3D wire trajectory. The Frenet Frame \cite{piccinelli2009framework} is the commonest way to provide a coordinate system at each point of a 3D line. Each image is registered to the corresponding point on the 3D wire and is perpendicular to it so that the tangent vector is normal to the image plane. Given the roto-translation matrices $R_1$, $R_2$, ... , $R_n$ where $n$ is the total number of OCT slices, the strut points cloud is registered along the 3D wire path. More precisely the set of $R_i$ is applied to 3D point cloud frame by frame:

$$
S_{cur,i}=R_i S_i,
$$
where $S_i$ stores the strut points of the $i$th OCT frame and $S_{cur,i}$ is the same set of points after the transformation. The labeling in rings and beam performed in the previous section is straightforwardly inherited from the registered point cloud. The latter becomes the input for the 3D volumetric reconstruction.

\subsection{Stent Volumetric Reconstruction}
Since the OCT was taken immediately after the stent deployment, the absorption process has not started yet. Hence, ring cross-section has dimension $ 150 \mu m \times 150 \mu m$ while the beam has dimension $ 150 \mu m \times 200 \mu m$, as for the Abbott Bioresorbable stent design. The points of the skeleton are regarded as the barycenter of each section. To each point we associate the intrinsic Frenet frame of reference, where one vector is tangent to the spline, the other two ({\it normal} and {\it bi-normal} vector) \cite{piccinelli2009framework} belong to the plane normal to the spline. Since the relative rotation of the rectangular section of the strut is unknown, we assume that the internal and external edge of the strut cross-section is orthogonal to the radial vector pointing from the lumen center to the strut barycenter (centroid), and the cross-sections in the deployed configuration are reconstructed accordingly. For each point of the spline interpolating the barycenters, the radial vector $r$ is computed and the four vertexes of the cross-section are collocated at the proper distance from the center with two edges orthogonal  (internal and external) and two edges parallel to $r$ (top and bottom), respectively, as shown in Fig. \ref{fig_5l}. At the end of this step, for each ring or beam component of the skeleton, an ordered list with the coordinates of the four vertexes of each cross-section is computed. This procedure relies on the following steps.

\subsubsection{Wireframe Reconstruction (Skeletonization).}
We construct the stent skeleton by a composite interpolation of the labeled centroids, where the different segments correspond to the different stent components. Centroids of each ring/beam are interpolated by a cubic spline. In fact, cubic splines minimize the elastic energy of a continuum passing through the interpolation point, a circumstance that yields an excellent consistency to the physical problem \cite{de1978practical}. The procedure creates a line in space, so it is necessary to provide for each spline a parametric representation in the form $[x(s), y(s), z(s)]$, where $s$ is the parameter. The convenient parameterization is different for beams and rings as they have a different location in space. Rings are circular frames properly parameterized by the polar coordinate $\vartheta$ introduced above and to be interpolated by periodic splines. Beams can be parameterized by the axial coordinate $z$ to be interpolated by natural splines. However, the ring splines may occasionally fail for that the noise affecting the data leads to un-smooth and nonrealistic wiggles (Fig. \ref{fig_7l}). Hence we resort to a modified parametrization called Lee's Centripetal Scheme \cite{lee1989choosing}, which always gives smooth spline interpolation for the rings. 

The most critical step in this stage is the identification of the junction between a ring and its beams. If a junction point has already been identified in the previous analysis, then the connection is trivial. On the contrary, when two adjacent components (one ring and one beam) do not have a common point, the beam is connected to the closest location on the ring spline. 

In summary: (1) we compute the periodic splines for the rings; (2) we compute the missing intersections between each ring and the corresponding beam; (3) we compute the beam natural spline interpolation. The results of a pair of rings and corresponding connectors are illustrated in Fig. \ref{fig_7l}.

\subsubsection{Surface Reconstruction.}
\begin{figure*}[htbp]
	\centering
	\subfloat[Skeletonization]{\includegraphics[width=3.5in,height=2.9in]{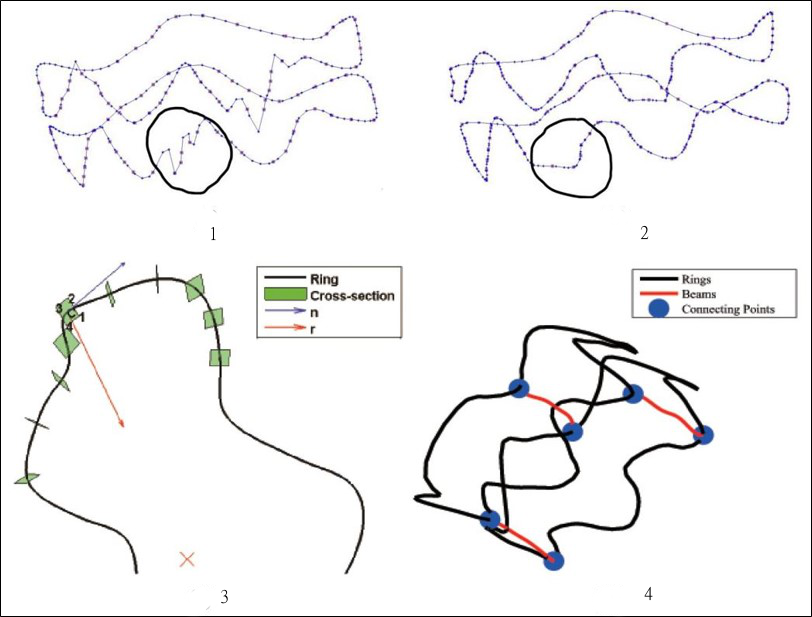}%
		\label{fig_7l}}
	\hfil
	\subfloat[Joint Reconstruction]{\includegraphics[width=3.5in,height=1.6in]{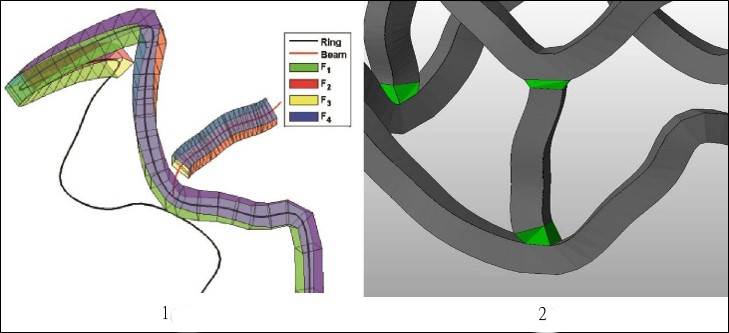}%
		\label{fig_7r}}
	\caption{(a) Skeletonization: (1) Polar coordinate $\vartheta$ are the parameter for the rings. Local oscillations of the interpolation generate wiggles. (2) 
		Eugene Lee's centripetal scheme smoothens the oscillations. 
		(3) Samples of registered and re-oriented cross-sections along the ring wireframe. (4) Connected stent wire-frame.
		(b) Joint Reconstruction: (1) Sample of bounding facets assembled on a ring and a beam. (2) The joints between rings and beams are constructed (green regions).}
	\label{fig_7}
\end{figure*}

As previously noted, the skeleton lines are regarded as the location of the centers of cross-sections of the 3D stent. We place 100 cross sections along each continuous skeleton lines, triangulate the resulting four lateral surfaces and save the results as a surface triangulation (STL) file. For each ring, this directly results in a closed surface. For the beams, we first construct open surfaces that end with no intersection with the rings - Fig. \ref{fig_7r}. Successively, the closest vertexes on the rings are selected to construct the intersection with the beams. To create a smooth connection between beam and ring, a rectangular hole is created on the ring structure in the proximity of the beam's end. The two open sections are then connected with two additional triangles per lateral surface. The surface reconstruction on the stent skeleton is automatic, and the reconstruction application is written in Matlab R2015a (MathWorks Inc., Natick, MA, 01760)

\section{Results}
In this section, we present the results of reconstructions from the clinical OCT images and the accuracy of the validation using the virtual stent geometry. 

\subsection{Clinical Cases}
\begin{figure*}[htbp]
	\centering
	\includegraphics[width=4.7in]{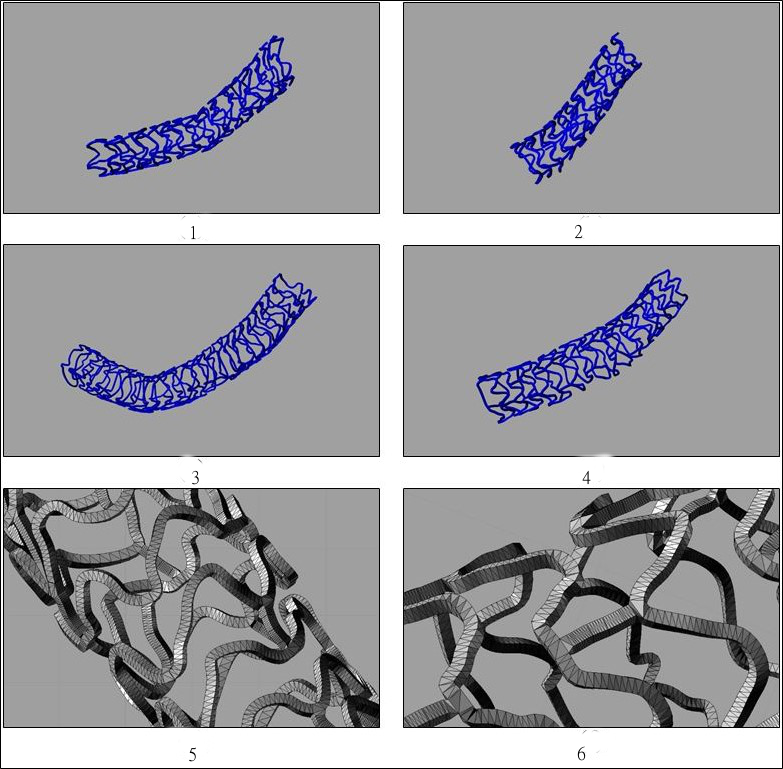}
	\caption{Reconstructed stent geometry of selected cases: (1) Case 1, (2) Case 2, (3) Case 3, (4) Case 4 . Panels (5) and (6) are zoomed-in pictures of Case 12.}
	\label{fig_8l}
\end{figure*}

In Fig. \ref{fig_8l}, we display the reconstructed patient-specific stents for the four selected cases (Case 2, 3, 6 and 12). These cases are selected out of the pool of 16 patients already reconstructed and analyzed with CFD - see \cite{gogas2013biomechanical,gogas2015computational,gogas2016novel} - as they cover the different typologies of patients we have encountered so far.  

\begin{figure*}[htbp]
	\centering
	\includegraphics[width=4.7in]{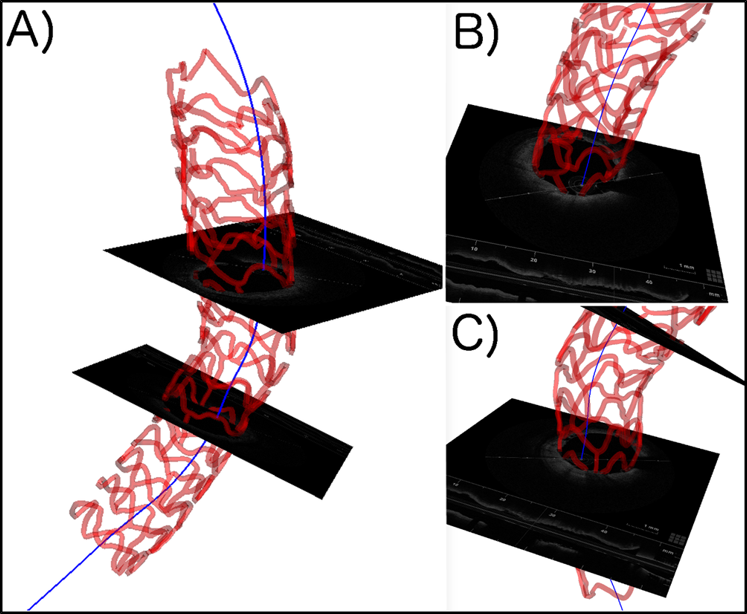}
	\caption{3D representation of the Patient-specific stent following the true vessel anatomy as defined by angio-derived OCT wire (blue line).}
	\label{fig_8r}
\end{figure*}

In Fig. \ref{fig_8r}, the registered OCT images are overlapped to the 3D reconstructed stent for a visual inspection of the correspondence between the detected struts and the images. As shown in Fig. \ref{fig_8r}, the reconstructed stent geometry is in good agreement with the patient's OCT images as the reconstructed 3D struts go through the 2D strut boxes on the OCT images. The automatic (i.e. before manual patching)  strut detection accuracy for the selected 4 cases ranges from 80\% to 85\% (Table \ref{tab: 1}). The stent reconstruction method made the semi-automatic patient-specific stent reconstruction possible within a reasonable time frame (Table \ref{tab: 2}). We report the average time cost to reconstruct one stent that covers at least 120 OCT frames.

\begin{table}


	\caption{Statistics of the Strut Detection Procedure on Clinical Cases}

	\label{tab: 1}       % Give a unique label


	\begin{tabular}{lllll}

		\hline\noalign{\smallskip}

		Case &2&3&6&12\\

		\noalign{\smallskip}\hline\noalign{\smallskip}

		No. of Frames &100&63&149&140\\

		Automatically Detected. &783&602&1152&1305\\

		False Positives. &18&34&18&55\\

		%Correctly detected struts&765&568&1134&1250\\[1mm]

		Manually Patched &180&101&251&215\\

		Total  &945&669&1385&1465\\

		\% Detected.  &80.9&84.9&81.9&85.3\\

		\noalign{\smallskip}\hline

	\end{tabular}

\end{table}

\begin{table}


	\caption{Timelines of the Stent Reconstruction (Average)}

	\label{tab: 2}       % Give a unique label


	\begin{tabular}{ll}

		\hline\noalign{\smallskip}

		%Frames &100&63&149&140\\

		Step    & Time (Min) \\

		\noalign{\smallskip}\hline\noalign{\smallskip}

		Automatic Strut Detection &120 \\

		Manual Correct and Patching &20\\

		Flattening &10\\

		Pattern Recognition &60\\

		Correction &30\\

		3D Reconstruction &10\\

		Geometry Repair &30\\

		Other &    20\\

		Total &    300 \\

		\noalign{\smallskip}\hline

	\end{tabular}

\end{table}

\subsection{Phantom Analysis}
The stent reconstruction procedure was applied to the phantom stent described in the Material section with two different centerline spacing, $0.2\ mm$ and $0.1\ mm$, respectively. Centerline spacing is the distance between two consecutive phantom OCT slices. As mentioned earlier, the phantom is subject to bending and twisting to mimic possible distortions
occurring during deployment. The spacing of 0.2 and 0.1 $mm$ is consistent with our devices.

\begin{figure*}[htbp]
	\centering
	\includegraphics[width=3.5in]{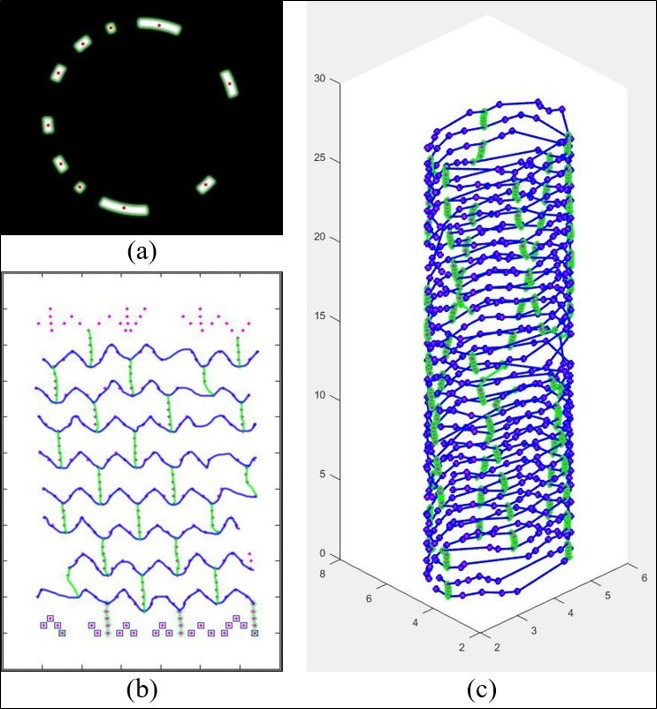}
	\caption{Phantom Data and Skeleton: (a) Detected struts from a Phantom OCT image. (b) Interpretive drawing of on 2D point cloud of the phantom stent. (c) Connected 3D point cloud of the phantom point cloud. }
	\label{fig_9l}
\end{figure*}

\begin{figure*}[htbp]
	\centering
	\subfloat[Phantom Reconstruction]{\includegraphics[width=3.5in,height=3.5in]{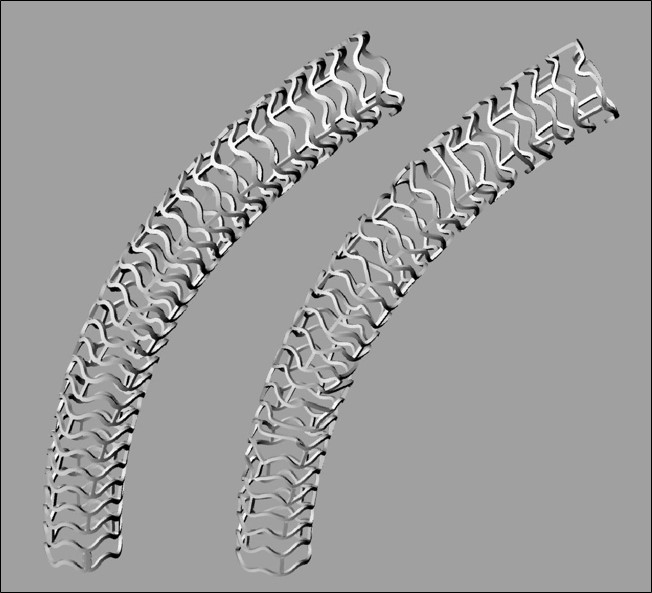}%
		\label{fig_9c}}
	\hfil
	\subfloat[Phantom Validation]{\includegraphics[width=3.5in,height=3.5in]{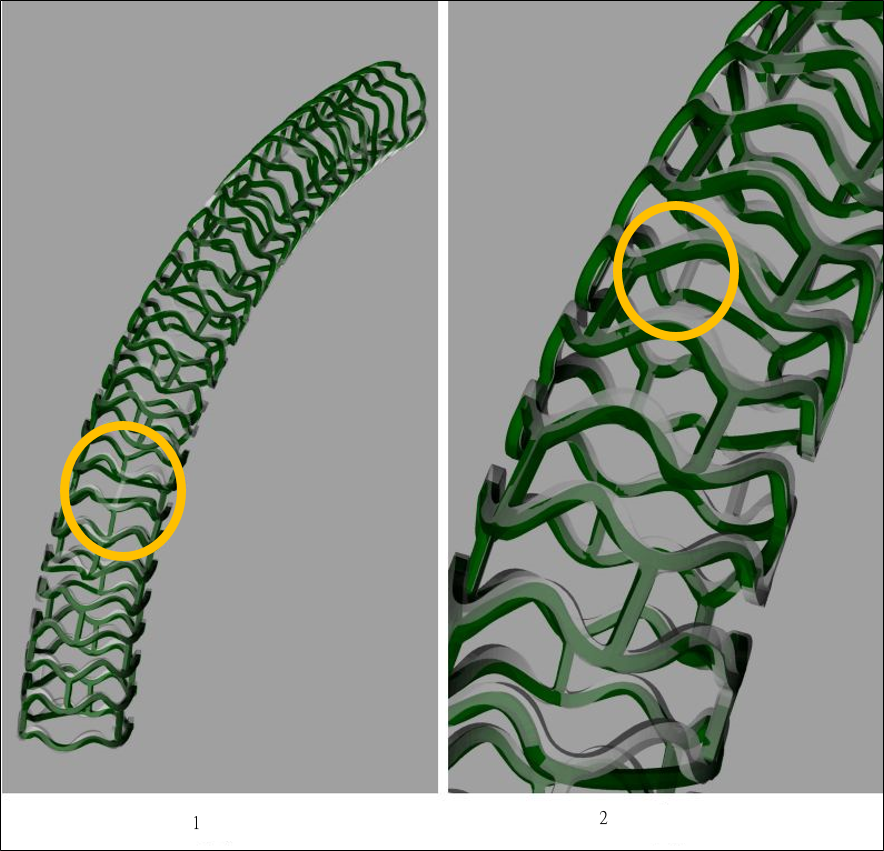}%
		\label{fig_9r}}
	\caption{(a) Phantom Reconstruction: Phantom stent (left) and Reconstructed stent (right). (b) Phantom Validation: (1) Reconstructed stent in green and phantom stent in transparent gray. (2) Zoomed-in picture (C)(1). The circles identify a beam structure not reconstructed because it is completely hidden by the catheter shadow.}
	\label{fig_9}
\end{figure*}

The phantom stent and the reconstructed stent using $0.2\ mm$ are displayed in Fig. \ref{fig_9c} side by side to give a visual inspection of the correspondence between the two models in the worst case scenario (0.2 vs. 0.1 cases). The reconstruction indeed captured the topological shape of the rings and the beams and the curvature of the original design.

We assess quantitatively in two aspects, the volume of the reconstructed stents and their physical position by measuring the overlapping between the phantom and its reconstruction.  The exact volume of the phantom is $V_p= 11.8789\ mm^3$. With spacing $0.2\ mm$, the reconstructed model finds a volume $V_{r,0.2}=11.3876\ mm^3$; the overlapping volume is $V_{o,0.2}=6.6408\ mm^3$. These numbers significantly improve with a spacing of 0.1 $mm$. In fact, in this case we have $V_{r,0.1}=11.8793\ mm^3$, and the overlapping volume increases to $V_{o,0.1}=9.0764\ mm^3$.

We summarize the performances with two indexes, volumetric accuracy $VA$ and Positional Accuracy $PA$: 
$$ VA = \frac{V_r}{V_p} (100\%) \qquad PA = \frac{V_o}{V_p} (100\%), $$
where $V_r$ is reconstructed volume, $V_p$ is phantom stent volume, and $V_o$ is overlapping volume. Table \ref{tab: 3} details the results. The reconstructed stent has high volumetric accuracy.  Not surprisingly, the positional accuracy is lower, showing that even small distortions that lead to an overall correct evaluation of the volume may generate positional errors. It is worth stressing the significant improvement induced by the refinement of the inter-slice distance. Such improvement gives an idea of how the limitations of the imaging impact the accuracy. The wireframe is reconstructed by cubic splines calculated over the points reconstructed. As it is well known, error interpolation in splines decreases with the number of points available (see e.g. \cite{de1978practical}). The fact that adding more nodes with an intra-slice distance of 0.1 $mm$ significantly reduces the positional error suggests that the interpolation is the major reason for the mismatch of the results.  The distance of 0.2 $mm$ is our worst-case scenario, as in our clinical OCT images were collected with either 0.1 or 0.2 $mm$. It is then reasonable to argue that our positional accuracy ranges from $55\%$ to $75\%$, which is consistent with other results. In \cite{chiastra2015computational}, an error of $20 \%$ on overlapping area on each slice is calculated, in the framework of a virtual deployment procedure, while our error refers to the total volumetric positional accuracy (i.e., on the total volume of the stent).

We will investigate in the follow-up of the present work on how this error affects the reconstruction of the stented lumen and consequently the sensitivity on the computational hemodynamics. It should be noted that the quantitative assessment of the stent volume is extremely accurate.
The shorter inter-slice distance improves an estimation which is already accurate in the worst case scenario. This improvement confirms our assumptions that as OCT spacing decreases our reconstruction method can produce more accurate stent geometry.

\begin{table}
	\caption{Reconstruction Accuracy}
	\label{tab: 3}       % Give a unique label
	\begin{tabular}{lll}
		\hline\noalign{\smallskip}
		%Frames &100&63&149&140\\
		& Spacing $0.2\ mm$ & Spacing $0.1\ mm$ \\
		\noalign{\smallskip}\hline\noalign{\smallskip}
		$V_r$  & $11.3876\ mm^3$ & $11.8793\ mm^3$ \\
		VA & $95.86\%$ & $100\%$ \\
		
		$V_o$  & $6.6408\ mm^3$ & $9.0764\ mm^3$ \\
		PA & $55.90\%$ & $76.41\%$ \\
		\noalign{\smallskip}\hline
	\end{tabular}
\end{table}

\section{Discussion and Limitations}
To the best of our knowledge, the workflow presented here is the first contribution toward the systematic patient-specific reconstruction of BVS required by Computer Aided Clinical Trials, covering a significant number of patients. The procedure features the following peculiarities.

\paragraph {Multimodal Imaging} OCT images are the core of the imaging procedure for the spatial resolution. However, OCT is not enough for a complete reconstruction, since the intravascular view alone does not give the location of the coronary and the stent in space. Registration with other images is required. Different from \cite{chiastra2015computational} where CT images are used, we use bi-planar angiographies to extract the curved centerlines, as these are part of the clinical routine in our group.
    
\paragraph{Semi-automated stent detection} On OCT slices strut detection is partially automated to face a large number of patients (each one featuring many slices). Empirical criteria have been developed for this purpose. However, the variety of post-deployment configurations still requires a manual surveillance. For instance, malapposed stents may be erroneously filtered by empirical criteria requiring the struts to be close enough to the vessel wall. Automated selection of optimal values of the processing parameters like $\gamma$  is an interesting follow-up of the present work. 

\paragraph {Stent skeletonization} The stent skeletonization is based on interpolation procedures. Even though these are mathematically rigorous and well rooted in the theory, they empirically fill the lack of knowledge due to the finite resolution and the presence of the catheter shadow.

\paragraph {Stent Volumetric Reconstruction} volumetric expansion relies on the assumption of no structural deformation of the stent. While this is in general not true, the mild dependence of the volumetric error on the inter-slice distance suggests that the impact of this assumption on the geometrical reconstruction and most importantly on the hemodynamics analysis is minor in comparison with the other sources of inaccuracy.

Results obtained on the phantom point out that the procedure yields clinically reliable results. The recovery of the exact original geometry is only partial for a series of approximation impacting the final result. (i) The major source of error is the finite axial resolution. Also, the pull-back maneuver is affected by rotational oscillations not detectable from the images. (ii) The catheter wire introduces a shadow that produces a lack of information in the images. The arc length of the hidden coronary (assuming a radius of $2~mm$ and for an angle of $\pi/6$) is about $1~mm$, vs. the edge of $0.15~mm$ of the beam. This shadow occurred in the phantom case, where the catheter completely masked 9 out of 56 beams. (iii) Multimodal image registration is affected by numerical errors. (iv) Interpolation and volumetric reconstruction both rely on empirical assumptions. 

Ideally, one could obtain the perfect geometry by simulating the physical deployment. This approach is in practice troublesome currently for lack of information on (i) the initial position of the undeployed stent, (ii) the force applied by the balloon, (iii) the material properties of the surrounding coronary arteries where the stent apposes to; and for the computational times required by this procedure. 

We intend to reduce the limitations and errors of the current procedure by an ``assimilation'' approach. The undeployed design of the stent can be used to correct current errors, by registering it to the current reconstruction. The comparison between the virtually deployed stent obtained by the registration and the reconstructed stent will help to reduce inaccuracies, for instance by filling the gaps induced by the catheter shadow. It is also expected to improve the automation of the procedure, as categorization will be performed directly by the registration. Mathematical foundations of our procedure for this step are reported in \cite{LRYV}.

Results obtained on the real cases demonstrate anyway that the procedure is affordable with reasonable timelines for clinical trials of the order of hundreds of patients followed up at different times. As a matter of fact, the semi-automated system is currently used on the ABSORB study within an established procedure. Extensive reconstruction of many patient-specific BVS stents is possible \cite{gogas2015computational,gogas2016novel}. For the purpose of assessing the interplay of hemodynamics and geometry, we argue that these results provide a significant step toward patient-specific clinically accurate geometrical reconstruction to be eventually finalized by CFD analysis. In fact, as for the preliminary results obtained so far (that will be presented in the follow-up of the present work), we guess that the geometrical errors do not prevent clinically reliable analysis over large pools of patients.

\section{Conclusions and Perspectives}
The inclusion of an accurate reconstruction of the patient-specific stent geometry is critical for assessing the impact of the presence of the struts on the local hemodynamics. The quantitative analysis of the potential sequence of events, triggered by the abnormal size of the struts, requires extensive investigations on statistically significant populations of patients to follow up in time. Numerical tools based on mathematical models of the hemodynamics are the method of choice, as they provide quantitative insights of relevant hemodynamics indexes. This workflow has tremendous potential for the ultimate assessment of pathological reactions. Nevertheless, it relies on many steps, the accurate reconstruction of patient-specific coronaries being one of the most important. Reliability and automation are critical features for CACT based analysis. 

The present work opens the path to systematic quantitative analysis based on CFD, by presenting a pipeline that assembles data (multimodal images) and geometrical primitives for a 3D characterization of the post-deployed stents. In the follow-ups of the present work, we will present the methodological steps that are required to run fluid dynamics simulations. In particular, the extraction of the stented lumen and its reticulation. In clinically oriented works,  we present the results obtained by extensive simulations, pointing out if and how the struts may trigger anomalous blood flow patterns, by extensive quantitative analysis of the Wall Shear Stress around the struts \cite{gogas2013biomechanical,gogas2015computational,gogas2016novel}.

At a methodological level, a natural question arises when extending the present approach to different stents, in particular, metallic ones. While the core of the presented pipeline remains valid, specific adjustments will be required for the proper detection of struts with different size and different impact on the images. 

\begin{acknowledgements}

The authors would like to thank Abbott Laboratory for the imaging data and research funding.
\end{acknowledgements}

\bibliographystyle{unsrt}
\bibliography{MBECbib.bib}  % name your BibTeX data base

\end{document}